\documentclass[11pt,a4paper]{article}
\usepackage[hyperref]{acl2020}
\usepackage{times}
\usepackage{latexsym}

\usepackage{microtype}

\usepackage{graphicx}
\usepackage{url}
\usepackage{bm}
\usepackage{amssymb,amsfonts,amsmath}
\usepackage{algorithm}
\usepackage{algorithmic}
\usepackage{mathtools}
\usepackage{booktabs}
\usepackage{subcaption}
\usepackage{xspace}
\usepackage{dcolumn}
\usepackage{comment}
\usepackage{multirow}
\usepackage{multicol}
\usepackage{enumitem}
\usepackage{mathrsfs}
\usepackage{makecell}
\DeclareMathOperator*{\argmax}{arg\,max}
\DeclareMathOperator*{\argmin}{arg\,min}

\makeatletter
\newcommand*\bigcdot{\mathpalette\bigcdot@{.5}}
\newcommand*\bigcdot@[2]{\mathbin{\vcenter{\hbox{\scalebox{#2}{$\m@th#1\bullet$}}}}}
  
\makeatother
\aclfinalcopy 
\def\aclpaperid{3355} 

\title{Memory-Efficient Differentiable Transformer Architecture Search}
 
\author{Yuekai Zhao$^\dag$,~~Li Dong$^\ddag$,~~Yelong Shen$^\ddag$,~~Zhihua Zhang$^{\S}$,~~Furu Wei$^{\ddag}$,~~Weizhu Chen$^{\ddag}$\\
$^\dag$Academy for Advanced Interdisciplinary Studies, Peking University \\
$^\S$School of Mathematical Sciences, Peking University \\
$^\ddag$Microsoft Corporation \\
\texttt{\{yuekaizhao@,zhzhang@math.\}pku.edu.cn} \\
\texttt{\{lidong1,yeshe,fuwei,wzchen\}@microsoft.com} \\}

\date{}
\begin{document}
\maketitle
\begin{abstract}
Differentiable architecture search (DARTS) is successfully applied in many vision tasks. However, directly using DARTS for Transformers is memory-intensive, which renders the search process infeasible. To this end, we propose a multi-split reversible network and combine it with DARTS. Specifically, we devise a backpropagation-with-reconstruction algorithm so that we only need to store the last layer's outputs. By relieving the memory burden for DARTS, it allows us to search with larger hidden size and more candidate operations. We evaluate the searched architecture on three sequence-to-sequence datasets, i.e., WMT'14 English-German, WMT'14 English-French, and WMT'14 English-Czech. Experimental results show that our network consistently outperforms standard Transformers across the tasks. Moreover, our method compares favorably with big-size Evolved Transformers, reducing search computation by an order of magnitude.
\end{abstract}

\section{Introduction}
Current neural architecture search (NAS) studies have produced models that surpass the performance of those designed by humans~\citep{real2019regularized,lu2020neural}. 
For sequence tasks, efforts are made in reinforcement learning-based~\citep{pham2018efficient} and evolution-based~\citep{so2019evolved,wangetal2020hat} methods, which suffer from the huge computational cost. Instead, gradient-based methods~\citep{liu2018darts,jiang2019improved,yang2020ista} are less demanding in computing resources and easy to implement, attracting many attentions recently.

\begin{figure} 
    \centering
    \includegraphics[width=0.90\linewidth]{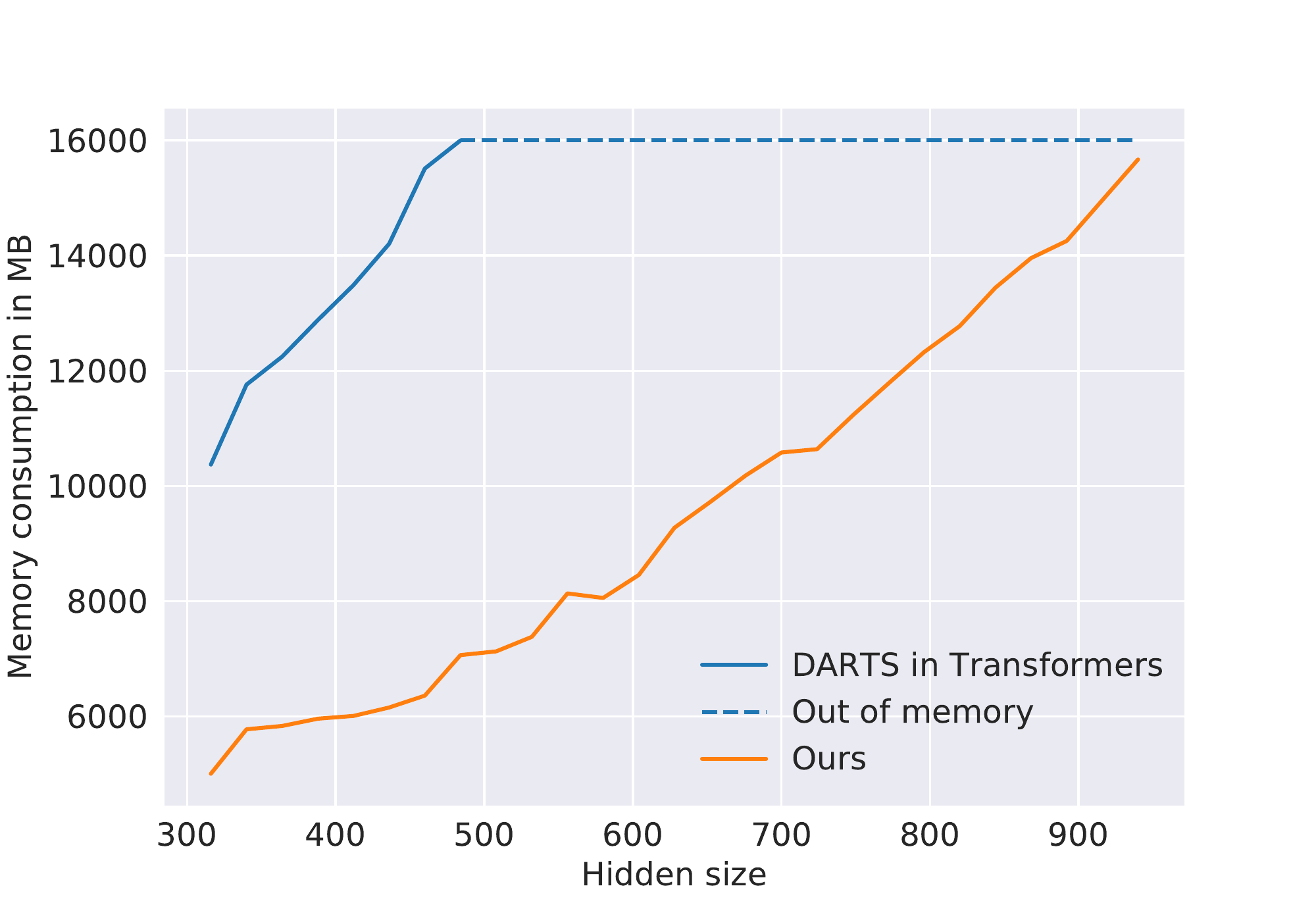}
    \caption{Memory comparison of using our reversible networks and Transformers as the backbone model of DARTS. Experiments are run on a single step of forward-backward pass on a batch of 3584 tokens with a NVIDIA P100 GPU.  Limited by GPU memory, DARTS in Transformers has to search in small sizes while evaluating in large sizes, which will cause performance gaps \citep{chen2019progressive}.} \label{fig:MemCom}
\end{figure}

The idea of gradient-based NAS is to train a super network covering all candidate operations. Different sub-graphs of the super network form the search space. To find a well-performing sub-graph, \citet{liu2018darts} (DARTS) introduced search parameters jointly optimized with the network weights. Operations corresponding to the largest search parameters are kept for each intermediate node after searching. A limitation of DARTS is its memory inefficiency because it needs to store the intermediate outputs from all its candidate operations. This is much more pronounced when we apply Transformers~\citep{vaswani2017attention} as the backbone of DARTS (the operation set is detailed in Section \ref{sect:Inst}). As shown in Figure \ref{fig:MemCom}, memory consumption grows extremely fast as we increase the hidden size $d$, quickly running out of memory as $d > 400$. As a result, we can only use a limited operation set or a small hidden size, which may lead to worse model performance. 

To address the unfavorable memory consumption issue in DARTS, we propose a variant of reversible networks. Each input of a reversible network layer can be reconstructed from its outputs. Thus, it is unnecessary to store intermediate outputs except for the last layer because we can reconstruct them during backpropagation (BP). Inspired by the idea of RevNets~\citep{gomez2017reversible}, we devise a multi-split reversible network. Each split contains a mixed operation search node to enable DARTS. Also, only a small modification of BP is needed to enable gradient calculation with input reconstruction. We show the memory consumption of our method in Figure \ref{fig:MemCom}, which on average halves the amount of memory required in the vanilla DARTS. We can search larger, deeper networks with a richer candidate operation set under the same memory constraint.

Our method is generic to handle various network structures. In this work, we focus on the sequence-to-sequence task.
We first perform the architecture search using the WMT'14 English-German translation task. The resulting architecture is then re-trained on three datasets: WMT'14 English-German, WMT'14 English-French, and WMT'14 English-Czech. We achieve consistent improvement over standard Transformers in all tasks. At a medium model size, we can have the same translation quality as the original ``big'' Transformer with 69\% fewer parameters. At a big model size, we exceed the performance of the Evolved Transformer~\citep{so2019evolved}, with the computational cost lowered by an order of magnitude. We will make our code and models publicly available.

\begin{figure*} 
    \centering
    \includegraphics[width=0.93\linewidth]{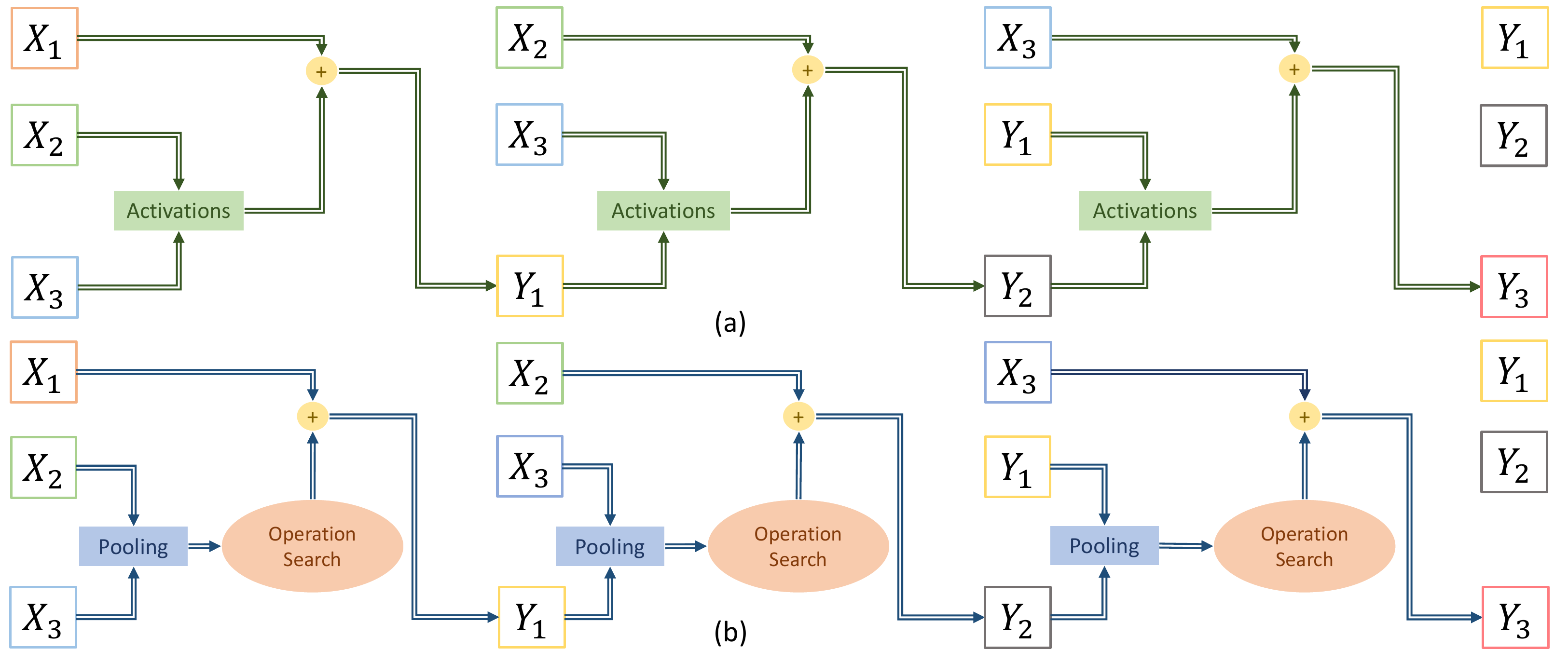}
    \caption{This figure is a demonstration of DARTSformer. (a) shows an $n$-split ($n=3$) reversible network, which serves the backbone of our method. (b) shows the design of activations to enable differentiable architecture search. Each $X_k$ and $Y_k$ are in $R^{l \times d}$. The $k$-th pooling takes the concatenation of $X_{i>k}$ and $Y_{i<k}$ as the input, and outputs a tensor in $R^{l \times d}$. The operation search gives a weighted average of the outputs of each candidate operation.} \label{fig:RevDarts}
\end{figure*}

\section{Methodology}
We give a detailed description of our method. In Section \ref{sect:DARTSinTrans}, we introduce DARTS and its memory inefficiency when applying in Transformers. In Section \ref{sect:MultiRev}, we propose a multi-split reversible network, which works as the backbone of our memory-efficient architecture search approach. Section \ref{sect:BPWithRecons} shows a backpropagation-with-reconstruction algorithm. In Section \ref{sect:MutliRevDARTS}, we manage to combine DARTS with our reversible networks. Finally, in Section \ref{sect:Inst}, we summarize the proposed algorithms with more details.

\subsection{Differentiable Architecture Search in Transformers} \label{sect:DARTSinTrans}
Following \citep{liu2018darts}, we explain the idea of differentiable architecture search (DARTS) within a one-layer block. Let $\mathscr{O}$ be the candidate operation set (e.g., Self Attention, FFN, Zero). Each operation $o \in \mathscr{O}$ represents some function that can be applied to the layer inputs or hidden states (denoted $X$). The key of DARTS is to use a mixed operation search node $f(X)$ to relax the categorical choice of a specific operation to a softmax over all candidate operations:
\begin{equation} \label{eq:OpSearch}
f(X) = \sum_{o \in \mathscr{O}} \frac{\text{exp}(\alpha_o)}{\sum_{o^{'} \in \mathscr{O}} \text{exp}(\alpha_{o^{'}})} o(X),
\end{equation}
where the $\alpha$ are trainable parameters of size $\vert \mathscr{O} \vert$ that determines the mixing weights. During searching, a one-layer block contains several search nodes. The task is to find a suitable set of $\alpha$ for each search node. At the end of the search, the resulting operation in each node is determined by:
\begin{align} \label{eq:Discretize}
    f = \argmax_{o \in \mathscr{O}} \alpha_o.
\end{align}
We optimize the $\alpha$ together with network weights $\theta$ by gradient descent. A good architecture means performing well on the searching validation set, such that we optimize $\alpha$ with validation loss $L_{val}$ and $\theta$ with training loss $L_{train}$:
\begin{align*}
    \begin{split}
        &\min_{\alpha} L_{val}(\theta^*(\alpha), \alpha), \\
        &\text{s.t.}\; \; \theta^*(\alpha) = \argmin_{\theta} L_{train}(\theta, \alpha).
    \end{split}
\end{align*}
In practice, we update $\alpha$ by $\nabla_{\alpha} L_{val}$ and $\theta$ by $\nabla_{\theta} L_{train}$ in each step. 

It is easy to directly apply DARTS in Transformers by replacing some or all operations in a Transformer block with mixed operation search nodes. For example, we can change the transformer decoder block from $\text{Self Attn} \rightarrow \text{Cross Attn} \rightarrow \text{FFN}$
to $\text{Search Node 1} \rightarrow \text{Cross Attn} \rightarrow \text{Search Node 2}$. Note that a search node outputs a weighted sum of different operations. To enable gradient calculation in the backward pass, we need to store every operation's output, which results in a steep rise in memory consumption during searching. Figure \ref{fig:MemCom} shows the memory consumption of using 2 search nodes in both Transformer encoder and decoder. DARTS run out of memory easily, even at a small hidden size. 

\subsection{Multi-split Reversible Networks} \label{sect:MultiRev}
To relieve the memory burden of DARTS in Transformers, we use reversible networks. A reversible network layer's input can be reconstructed from its output. Suppose a network is comprised of several reversible layers. We do not need to store intermediate outputs except the last layer, because we can reconstruct them from top to bottom during backpropagation (BP). Denote by $X$ and $f(X)$ the layer input and the layer output, respectively. $X$ is first split along the embedding/channel dimension into $n$ equal parts $\{X_1, \cdots, X_n\}$. A RevNets \citep{gomez2017reversible} alike operation is applied to each $X_k$, which yields $Y_k$. $f(X)$ is a concatenation of $\{Y_1, \cdots, Y_n\}$ along the split dimension:
\begin{align} \label{eq:MultiRevNets}
    \begin{split}
        &Y_1 = X_1 + G_1(X_{i>1}, \theta_1), \\
        &\ldots \\
        &Y_k = X_k + G_k(X_{i>k}, Y_{i<k}, \theta_k), \\
        &\ldots \\
        &Y_n = X_n + G_n(Y_{i<n}, \theta_n).
    \end{split}
\end{align}
$G_k$ is a mixed operation node during the architecture search process. After searching, $G_k$ is a deterministic operation given by $\argmax_{o \in \mathscr{O}} \alpha_o$. Detailed discussions can be found in Section \ref{sect:MutliRevDARTS}. 

The reversibility of Eq. \eqref{eq:MultiRevNets} needs rigorous validation, such that the input $X$ can be easily reconstructed from $f(X)$:
\begin{align} \label{eq:MultiRevNetsRecons}
    \begin{split}
        &X_n = Y_n - G_n(Y_{i<n}, \theta_n), \\
        &\ldots \\
        &X_k = Y_k - G_k(X_{i>k}, Y_{i<k}, \theta_k), \\
        &\ldots \\
        &X_1 = Y_1 - G_1(X_{i>1}, \theta_1). \\
    \end{split}
\end{align}
Part (a) of Figure \ref{fig:RevDarts} illustrates a 3-split reversible network, which we frequently employ throughout our experiments for simplicity.
\begin{algorithm}[t]
    \caption{BP-with-reconstruction Algorithm for Multi-Split Reversible Networks}
        \begin{algorithmic}[1] \label{algo:revbp}
        \REQUIRE ~~\\
            Layer output: $f(X) =  [Y_1, \cdots, Y_n] $; \\
            Total derivatives: $\textit{d}f(X) =  [\textit{d}Y_1, \cdots, \textit{d}Y_n]$; \\
            Operations: $G_1, \cdots, G_n$;
        \ENSURE ~~\\
                Layer input: $X = [X_1, \cdots, X_n]$; \\
                Derivatives of $X$: $\textit{d}X = [\textit{d}X_1, \cdots, \textit{d}X_n]$;
        \STATE $X = \{\}; \textit{d}X = \{\}; Y = \{Y_1, \cdots, Y_n\}$ \\
        \FOR{k in $n$ to $1$} 
        \STATE $C$ = $Y_k$; $Y = Y \setminus \{Y_k\}$ \\
        \IF{$k == n$}
        \STATE $\text{grad}_k = \textit{d}Y_k$
        \ELSE
        \STATE $\text{grad}_k = \textit{d}Y_k + C.\text{grad}$
        \ENDIF
        \STATE $g_k = G_k(X, Y, \theta_k)$; $g_k$.backward\footnotemark($\text{grad}_k$)
        \STATE $X_k = C - g_k; X = X \cup \{X_k\}$
        \ENDFOR
        \STATE $\textit{d}X_1 = \text{grad}_1, \textit{d}X = \{\textit{d}X_1\}$
        \FOR{k in $2$ to $n$}
        \STATE $\textit{d}X_k = X_k.\text{grad} + \text{grad}_k$
        \STATE $\textit{d}X = \textit{d}X \cup \{\textit{d}X_k\}$
        \ENDFOR
        \end{algorithmic}
\end{algorithm}
\subsection{Backpropagation with Reconstruction} \label{sect:BPWithRecons}
Consider the problem of backpropagating (BP) through a reversible layer. Based on the layer output $f(X) = \text{Concat}(Y_1, \cdots, Y_n)$ and its total derivative $\textit{d}f(X) = \text{Concat}(\textit{d}Y_1, \cdots, \textit{d}Y_n)$, we need to calculate the layer input $X = \text{Concat}(X_1, \cdots, X_n)$, its total derivative $\textit{d}X = \text{Concat}(\textit{d}X_1, \cdots, \textit{d}X_n)$, and the derivatives of the network weights $\textit{d}\theta_1, \cdots, \textit{d}\theta_n$.

We show the BP-with-reconstruction through a single layer in Algorithm \ref{algo:revbp}. $[\cdot]$ represents $\text{Concat}(\cdot)$ for simplicity reasons. In Line 9 of Algorithm \ref{algo:revbp}, $\textit{d}\theta_k$ is calculated as a side effect. Line 10 shows the reconstruction process, where each split $X_k$ is recovered in the order of $n$ to $1$. In Algorithm \ref{algo:revbp}, $\text{grad}_k$ works as a gradient accumulator, which keeps track of all derivatives associated with $X_k$. A repetitive application of Algorithm \ref{algo:revbp} enables us to backpropagate through a sequence of reversible layers. Only the top layer's outputs require storage, which makes it much more memory-efficient.

Roughly speaking, for a network with $N$ connections, the forward and backward passes require approximately $N$ and $2N$ add-multiply operations, respectively. Since we need to reconstruct $X$ from $f(X)$, the re-calculation requires another $N$ add-multiply operations, making it $33\%$ slower. Fortunately, we can only need Algorithm \ref{algo:revbp} for architecture search and will re-train the resulting network with ordinary BP. The search process turns out to converge fast. The computational overhead does not become a severe problem.
\footnotetext{Automatic differentiation routines, e.g. \texttt{tf.gradient}, \texttt{torch.autograd.backward}}

\subsection{DARTS with Multi-split Reversible Networks} \label{sect:MutliRevDARTS}
Performing DARTS based on $n$-split reversible networks only requires specifying each $G_k$ in Eq. \eqref{eq:MultiRevNets}. Suppose that each $X_k \in R^{l \times d_n}$ ($l$ is the sequence length and $d$ is the hidden size, $d_n=\frac{d}{n}$), and that each $Y_k$ has the same size as $X_k$. The input of $G_k$ contains $n{-}1$ tensors in $R^{l \times d_n}$. To enable element-wise addition with $X_k$, the output of $G_k$ must also be in $R^{l \times d_n}$. 

\begin{figure} 
    \centering
    \includegraphics[width=0.95\linewidth]{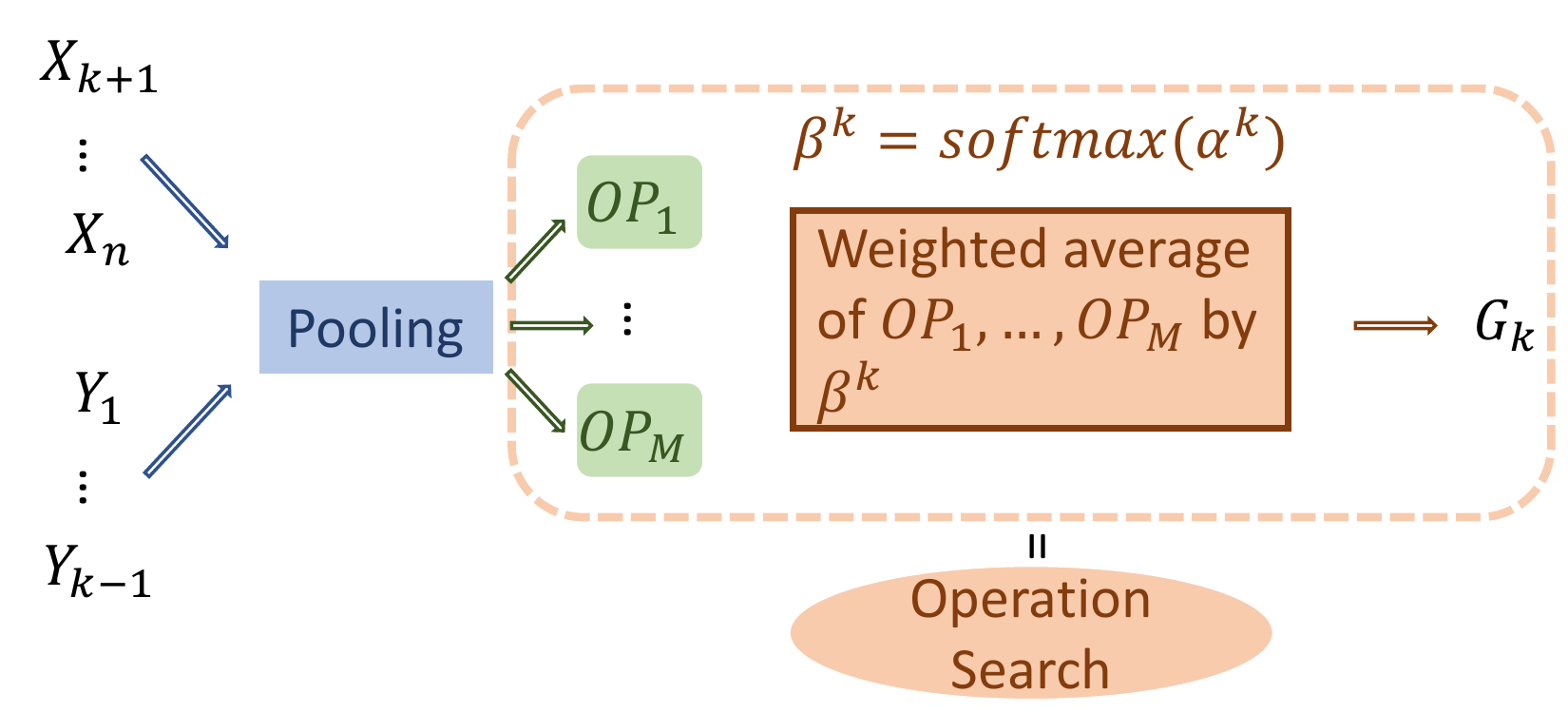}
    \caption{Pooling and operation search in each split.} \label{fig:GkFig}
\end{figure}

$G_k$ is factorized into two parts. The first part is a pooling operation, which takes an $l \times d_n \times (n-1)$ tensor as input, and outputs an $l \times d_n \times 1$ tensor. The second part is a mixed operation search node. $G_k$ is calculated as follows:
\begin{align} \label{eq:Gk}
\begin{split}
    &H_k = \text{Pooling}(X_{i>k}, Y_{i<k}), \\
    &G_k = \sum_{o \in \mathscr{O}} \frac{\text{exp}(\alpha^k_o)}{\sum_{o^{'} \in \mathscr{O}} \text{exp}(\alpha^k_{o^{'}})} o(H_k),
\end{split}
\end{align}
where $\alpha_k$ is randomly initialized.
Figure \ref{fig:GkFig} shows the design of $G_k$. By substituting each $G_k$ in Eq. \eqref{eq:MultiRevNets} with Eq. \eqref{eq:Gk}, we are able to use Algorithm \ref{algo:revbp} to perform memory-efficient DARTS. We call this method \emph{DARTSformer}, which is illustrated by Part (b) of Figure \ref{fig:RevDarts} in a 3-split case.

The overall search space size is critical to the performance of DARTSformer. In our experiments, we focus on sequence-to-sequence tasks where the encoder and the decoder are searched simultaneously. Suppose that we have an $m$-split encoder and an $n$-split decoder. We search $s$ consecutive layers. For example, $s=2$ means that we search within a 2-layer encoder block. Each layer in the block is an $m$-split reversible layer. The encoder contains several identical 2-layer blocks, the same to the decoder. The search space is of size $\vert \mathscr{O} \vert^{s(m+n)}$. If $\vert \mathscr{O} \vert$ is large, it can easily introduce a large search space even with small $m,n$ and $s$.

\subsection{Instantiation} \label{sect:Inst}
We describe the instantiation of DARTSformer in this section.
\paragraph{Operation Set}  The candidate operation set $\mathscr{O}$ is defined as follows:
\begin{align*}
    &\bigcdot \text{Standard Conv }w \times 1: \text{for }w \in \{3,5,7,11\}. \\
    &\bigcdot \text{Dynamic Conv }w \times 1: \text{for }w \in \{3,7,11,15\}. \\
    &\bigcdot \text{Self Attention.} \\
    &\bigcdot \text{Cross Attention: Only available to decoder.} \\
    &\bigcdot \text{Gated Linear Unit (GLU).} \\
    &\bigcdot \text{FFN.} \\
    &\bigcdot \text{Zero: Return a zero tensor of the input size.} \\
    &\bigcdot \text{Identity: Return the input.}
\end{align*}
The Dynamic Conv is from \citet{wu2019pay}. The Self Attention, Cross Attention and FFN are from \citet{vaswani2017attention}. We use 8 attention heads. The GLU is from \citet{dauphin2017language}.

Residual connections \citep{he2016deep} and layer normalization~\citep{jimmy16LN} are crucial for convergence in training Transformers~\citep{vaswani2017attention}. To make our network fully reversible, these two tricks can not be used directly. Instead, we put the residual connections and layer normalization within each operation $\tilde{o}(X) = \text{LayerNorm}(X + o(X))$, except for Zero and Identity.
\paragraph{Encoder and Decoder} We use an $n$-split encoder and an $(n{+}1)$-split decoder for DARTSformer. Each $G_k$ in the encoder takes the format of Eq. \eqref{eq:Gk}. Instead for the decoder, $G_{k<n+1}$ still follows Eq. \eqref{eq:Gk}, but the operation for the last split $G_{n+1}$ is fixed as Cross Attention. Our experiments show that this constraint on the decoder yields architectures with better performances. 
\paragraph{Search and Re-train}
\begin{algorithm}[t] 
    \caption{The framework of DARTSformer}
        \begin{algorithmic}[1] \label{algo:DARTSformer}
        \REQUIRE ~~\\
            Operation set: $\mathscr{O}$, Search parameters: $\alpha$; \\
            Network weights: $\theta$;
        \ENSURE ~~\\
            Best candidate network: $\mathcal{N}_{final}$;
        \STATE Setup a multi-split reversible network with operation search nodes $\mathcal{N}_{super}(\mathscr{O}, \alpha, \theta)$. \\
        \WHILE{$\alpha$ not converge}
        \STATE Update $\theta$ by Algorithm \ref{algo:revbp} with $L_{train}$. \\
        \STATE Update $\alpha$ by Algorithm \ref{algo:revbp} with $L_{val}$. \\
        \ENDWHILE
        \STATE Get $\mathcal{N}_{final}(\mathscr{O}, \alpha, \theta)$ with Eq. \eqref{eq:Discretize}. \\
        \end{algorithmic}
\end{algorithm}
We summarize the entire framework of DARTSformer in Algorithm \ref{algo:DARTSformer}. Note that the search process is the most memory intensive part, such that we use BP-with-reconstruction as shown in Line $2$-$5$ of Algorithm \ref{algo:DARTSformer}. 

\section{Experiment Setup}

\subsection{Datasets} \label{sect:dataset}
We use three standard datasets to perform our experiments as \citet{so2019evolved}: (1) WMT'18 English-German (En-De) without ParaCrawl, which consists of 4.5 million training sentence pairs. (2) WMT'14 French-English (En-Fr), which consists of 36 million training sentence pairs. (3) WMT'18 English-Czech (En-Cs), again without ParaCrawl, which consists of 15.8 million training sentence pairs. Tokenization is done by Moses\footnote{https://github.com/moses-smt/mosesdecoder}. We employ BPE~\citep{sennrichetal2016neural} to generate a shared vocabulary for each language pair. The BPE merge operation numbers are 32K (WMT'18 En-De), 40K (WMT'14 En-Fr), 32K (WMT'18 En-Cs). We discard sentences longer than 250 tokens. For the re-training validation set, we randomly choose 3300 sentence pairs from the training set. The evaluation metric is \textit{BLEU}~\citep{papinenietal2002bleu}. We use beam search for test sets with a beam size of 5, and we tune the length penalty parameter from $0.5$ to $1.0$. Suppose the input length is $m$, and the maximum output length is $1.2m + 10$.


\subsection{Search Configuration} \label{sect:SearchConf}
The architecture searches are all run on WMT'14 En-De. DARTS is a bilevel optimization process, which updates network weights $\theta$ on one dataset and search parameters $\alpha$ on another dataset. We split the 4.5 million sentence pairs into 2.5/2.0 million for $\theta$ and $\alpha$. Both $L_{train}$ and $L_{val}$ are cross entropy loss with a label smoothing factor of $0.1$. The split number $n$ is 2 for the encoder and 3 for the decoder. We set $s$ to $1$ or $2$, which means the super network contains several identical $1$-layer or $2$-layer blocks. The candidate operations are detailed in Section \ref{sect:Inst}, where $\vert \mathscr{O} \vert = 13/14$ for encoder and decoder, respectively. Along the analysis in Section \ref{sect:MutliRevDARTS}, the largest size of the search space is around 1 billion. We use a factorized word embedding matrix to save memory. $\vert V \vert$ is the vocabulary size, and $d$ is the hidden size. The original word embedding matrix $E \in R^{\vert V \vert \times d}$ is factorized into a multiplication of two matrices of size $\vert V \vert \times e$ and $e \times d$, where $e \ll d$. We let $e$ denote the embedding size. We set $e=256, d=960$. During searching, we set the dropout probability to $0.1$. Two Adam optimizers~\citep{AdamKingmaB14} are used for updating $\theta$ and $\alpha$, with $\beta_1 = 0.9$ and $\beta_2 = 0.98$. For $\theta$, we use the same learning rate scheduling strategy as done in \citet{vaswani2017attention} with a warmup step of $10000$. The maximum learning rate is set to $5 \times 10^{-4}$. For $\alpha$, we fix the learning rate to $3 \times 10^{-4}$ with a weight decay of $1 \times 10^{-3}$, which is the same as \citet{liu2018darts} does.

DARTSformer requires us to specify a pooling operation as stated in Eq. \eqref{eq:Gk}. We experiment with both max pooling and average pooling.
All searches run on the same 8 NVIDIA V100 hardware. We use a batch size of $5000$ tokens per GPU and save a checkpoint every 10,000 updates ($5000$ for $\theta$ and $5000$ for $\alpha$). Our search process finalizes after 60,000 updates. 

\subsection{Training Details}
All the networks derived from the saved checkpoints are re-trained on WMT'14 En-De to select the best performing one. We then train the selected network on all datasets in Section \ref{sect:dataset} to verify its generalization ability. We follow the settings of \citet{so2019evolved} with both a base model and a big model. For the base model, we still use $e=256, d=960$ without re-scaling. For the big model, we set $e=512, d=1824$. 
Unless otherwise stated, all the training run on 8 Tesla V100 GPU cards with the batch size of 5000 tokens per card.
\begin{table}[t]
    \centering
    \setlength{\tabcolsep}{1.1mm}
    \begin{tabular}{lccccc}
        \toprule[1.5pt]
         \textbf{Model} & \textbf{Pooling} & \thead{\textbf{Search} \textbf{$s$} \\ \textbf{Layers}} & \thead{\textbf{Model} \\ \textbf{Size}} & \textbf{BLEU} \\
        \midrule
        Transformer & - & - & 61.1M & 27.7\\
        ET & - & - & 64.1M & 28.2 \\ 
        \midrule
        Sampling & max & 2 & 60.1M & 18.7 \\
        Sampling & avg & 2 & 61.6M & 16.8 \\
        DARTSformer & max & 1 & 64.5M  & 27.9 \\
        DARTSformer & max & 2 & 65.2M & \textbf{28.4} \\
        DARTSformer & avg & 1 & 66.0M & 28.3 \\
        DARTSformer & avg & 2 & 63.4M & 28.3 \\
        \bottomrule[1.5pt]
    \end{tabular}
    \caption{BLEU scores of various search setups on WMT'14 En-De test set. ET is the Evolved Transformer~\cite{so2019evolved}. We use a $2$-split encoder and a $3$-split decoder.}
    \label{table:SearchRes}
\end{table}
\begin{table}[t]
    \centering
    \setlength{\tabcolsep}{1.3mm}
    \begin{tabular}{lccccc}
        \toprule[1.5pt]
         \textbf{Model} & \textbf{Pooling} & \textbf{Splits} & \textbf{BLEU} \\
        \midrule
        DARTSformer & max & 2,3 & \textbf{28.4} \\
        DARTSformer & max & 3,4 & 28.0 \\
        DARTSformer & max & 4,5 &  27.4 \\
        \midrule
        DARTSformer & avg & 2,3 & \textbf{28.3} \\
        DARTSformer & avg & 3,4 & 27.9 \\
        DARTSformer & avg & 4,5 & 27.1 \\
        \bottomrule[1.5pt]
    \end{tabular}
    \caption{BLEU scores of DARTSformer with different split numbers on WMT'14 En-De test set. We use an $n$-split encoder and an $n+1$-split decoder. We searching through 2 consecutive layers.}
    \label{table:SearchResSplit}
\end{table}
\begin{table}[t]
    \centering
    \setlength{\tabcolsep}{1.3mm}
    \begin{tabular}{lccc}
        \toprule[1.5pt]
         \textbf{Model} & \textbf{Price} & \textbf{Steps} & \textbf{Hardware} \\
        \midrule
        ET & \$150k & $4.2 \times 10^8$ & 200 TPUs \\ 
        DARTSformer & \$1.25k & $4.8 \times 10^5$ & 8 V100 \\
        \bottomrule[1.5pt]
    \end{tabular}
    \caption{Comparison for search cost between Evolved Transformer (ET; \citealt{so2019evolved}) and DARTSformer. The price for ET is from \citet{strubell2019energy}.}
    \label{table:CostComp}
\end{table}
\begin{figure}
    \centering
    \includegraphics[width=0.95\linewidth]{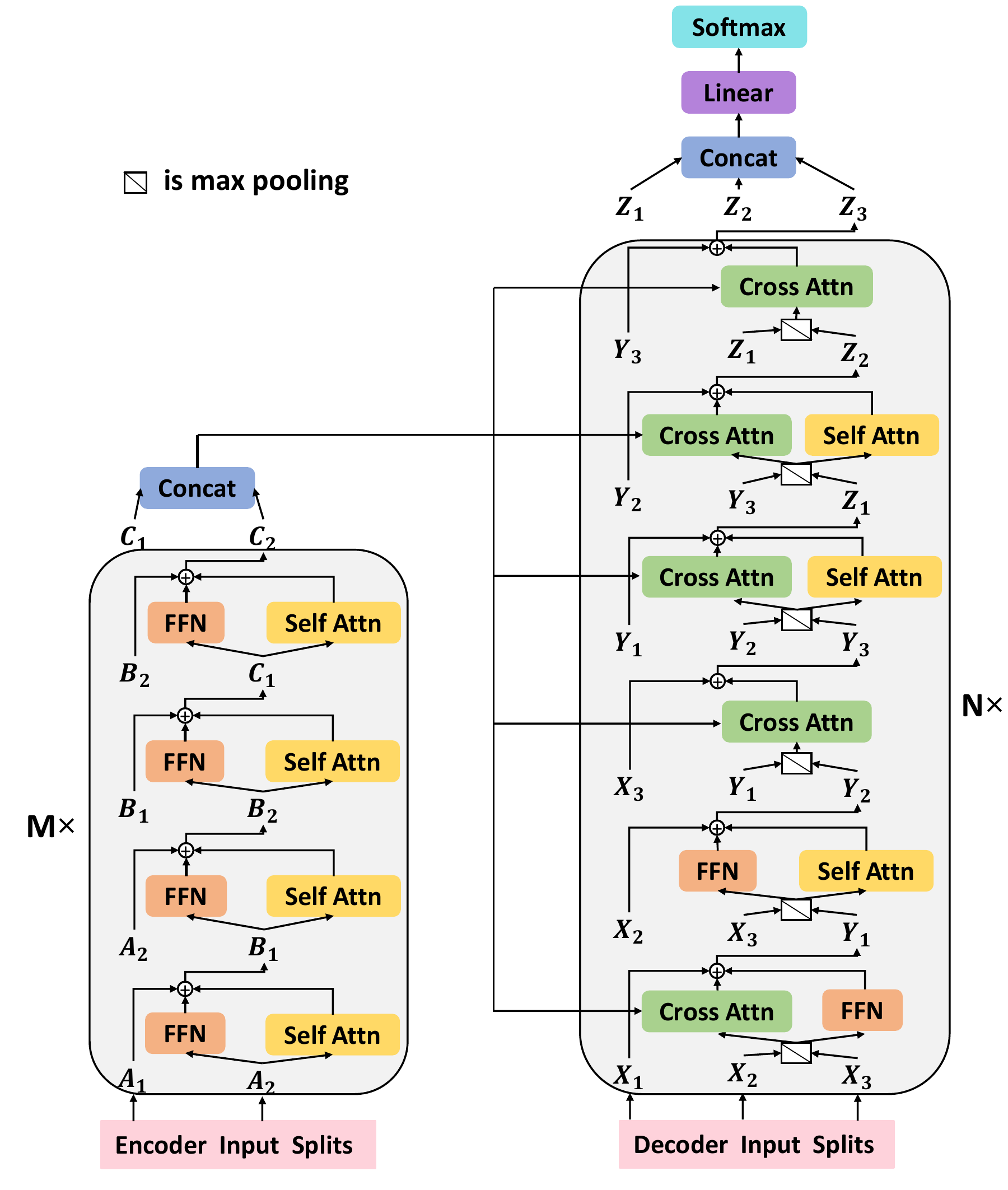}
    \caption{Architecture searched by DARTSformer.} \label{fig:DARTSformerArch}
\end{figure}
\section{Results}
\subsection{Comparison Between Search Setups} \label{sect:CompSetup}
We search through a different number of consecutive layers with different pooling operations. For re-training, we use the same learning rate scheduling strategy as in searching. We also keep the dropout rate unchanged. Results are summarized in Table \ref{table:SearchRes}. DARTSformers yields better results than standard Transformers in all experimental setups. The maximum performance gain is 0.7 BLEU with max pooling when searching through $2$ consecutive layers. Also, DARTSformer achieves slightly better results than the Evolved Transformer in three out of four runs. 

We compare the search cost between the Evolved Transformers and DARTSformer from various aspects. DARTSformer takes about 40 hours to run on an AWS p3dn.24xlarge node\footnote{https://aws.amazon.com/ec2/instance-types/p3/}. The price for a single run of search is about \$1.25k. As reported by \citet{strubell2019energy}, the search process of Evolved Transformer takes up to \$150k, which is extremely expensive. As for hardware, the evolutionary search employs 200 TPU V.2 chips to run, while our method only uses 8 NVIDIA V100 cards. The reason for the evolutionary search algorithm's huge cost is that it requires training multiple candidate networks from scratch. We compare the number of parameter update steps in Table \ref{table:CostComp}. The evolutionary search needs approximately 874 times more update steps than our method.

A simple sampling-based NAS method \citep{guo2020single} can also reduce memory consumption. For each batch of training data, we set $G_k$ in Eq. \eqref{eq:Gk} as a uniformly sampled operation from the candidate set $\mathscr{O}$. The search parameters $\alpha$ are discarded, and the resulting network is produced from an evolutionary search by evaluating on the re-training validation set. This method performs poorly in machine translation, as shown in Table \ref{table:SearchRes}. We find that sampling-based methods favor large-kernel convolutions and that the resulting architectures tend to generate repetitive sentences.

We also experiment with increased split numbers. As shown in Table \ref{table:SearchResSplit}, an increased split number hurts the translation performance. The best results are all achieved by the smallest split. Also, the search process is harder to converge as the search space becomes too large. The re-training and inference speed will slow down when increasing the split number because more recurrence are introduced in the calculation as shown in Eq. \eqref{eq:MultiRevNets}. 

In the following sections, we try the best search result (DARTSformer + search 2 layers + 2 split + max pooling) in various sequence-to-sequence tasks to see its generalization ability. We show this searched architecture in Figure \ref{fig:DARTSformerArch}.

\subsection{Performance of DARTSformer on Other Datasets} \label{sect:MoreRes}
First, we train DARTSformer with a base model size on three translation tasks in Section \ref{sect:dataset}. We would like to see whether DARTSformer only performs well on the task used for architecture search or generalizes to related tasks. Second, we scale up the model size and the batch size to see whether the performance gain of DARTSformer still exists. We compare DARTSformer with standard Transformers and Evolved Transformers with similar model sizes. Following \citet{vaswani2017attention}, the parameter size is around 62.5M/214.7M for the base model and big model, respectively. To match the settings of \citet{so2019evolved} when training big models, we increase the dropout rate to $0.3$ and the learning rate to $1 \times 10^{-3}$. We also accumulate gradients for two batches.

\begin{table}[t]
\subfloat[Comparison for Base Model Size]{
    \begin{tabular}{lccc}
    \toprule[1.5pt]
    Models & En-De & En-Fr & En-Cs \\
    \midrule
    Transformer & 27.7 & 40.0 & 27.0 \\
    ET~\cite{so2019evolved} & 28.2 & \textbf{40.6} & 27.6 \\
    DARTSformer & \textbf{28.4} & 40.1 & \textbf{27.9} \\
    \bottomrule[1.5pt]
    \end{tabular}
    }
    \quad
\subfloat[Comparison for Big Model Size]{
    \begin{tabular}{lccc}
    \toprule[1.5pt]
    Models & En-De & En-Fr & En-Cs \\
    \midrule
    Transformer & 29.1 & 41.2 & 28.1 \\
    ET~\cite{so2019evolved} & 29.3 & \textbf{41.3} & 28.2 \\
    DARTSformer & \textbf{29.8} & \textbf{41.3} & \textbf{28.5} \\
    \bottomrule[1.5pt]
    \end{tabular}
}
\caption{BLEU scores on WMT'14 translation tasks. ET is the Evolved Transformer. We use the best search result from different DARTSformer search setups.}
\label{table:MoreMTComp}
\end{table}

Results are shown in Table \ref{table:MoreMTComp}. At the base model size, DARTSformer steadily outperforms standard Transformers. We achieved the same translation quality (28.4 BLEU, reported by \citet{vaswani2017attention}) as the original big Transformer in WMT'14 En-De, with about 69\% fewer parameters. Also, the maximum BLEU gain is 0.9 in WMT'14 En-Cs, which is not the dataset we conduct our architecture search on. As for Evolved Transformers, we surpass their performance in two out of three datasets, and our search algorithm is more computationally efficient. At the big model size, DARTSformer exceeds both standard Transformers and Evolved Transformers, which indicates the good generalization ability of DARTSformer.

\subsection{Performance of DARTSformer vs. Parameter Size}
In Section \ref{sect:MoreRes}, DARTSformer consistently improves the performance with a model size comparable to the base and big Transformers. We are wondering whether the performance increase exists with smaller model sizes. We experiment with a spectrum of model sizes for standard Transformers and DARTSformer on WMT'14 En-De. Specifically, we use four embedding sizes for standard Transformers, [small:128, medium:256, base:512, big:1024], where its hidden size is identical to the embedding size. We also adjust the model size of DARTSformer accordingly. For base and big models, we use the results from Section \ref{sect:MoreRes}. For small and medium models, we set the learning rate to $5 \times 10^{-4}$, the dropout probability to $0.1$, and update the model parameters for 200,000 steps on the same 8 NVIDIA V100 hardware.

\begin{figure}
    \centering
    \includegraphics[width=0.96\linewidth]{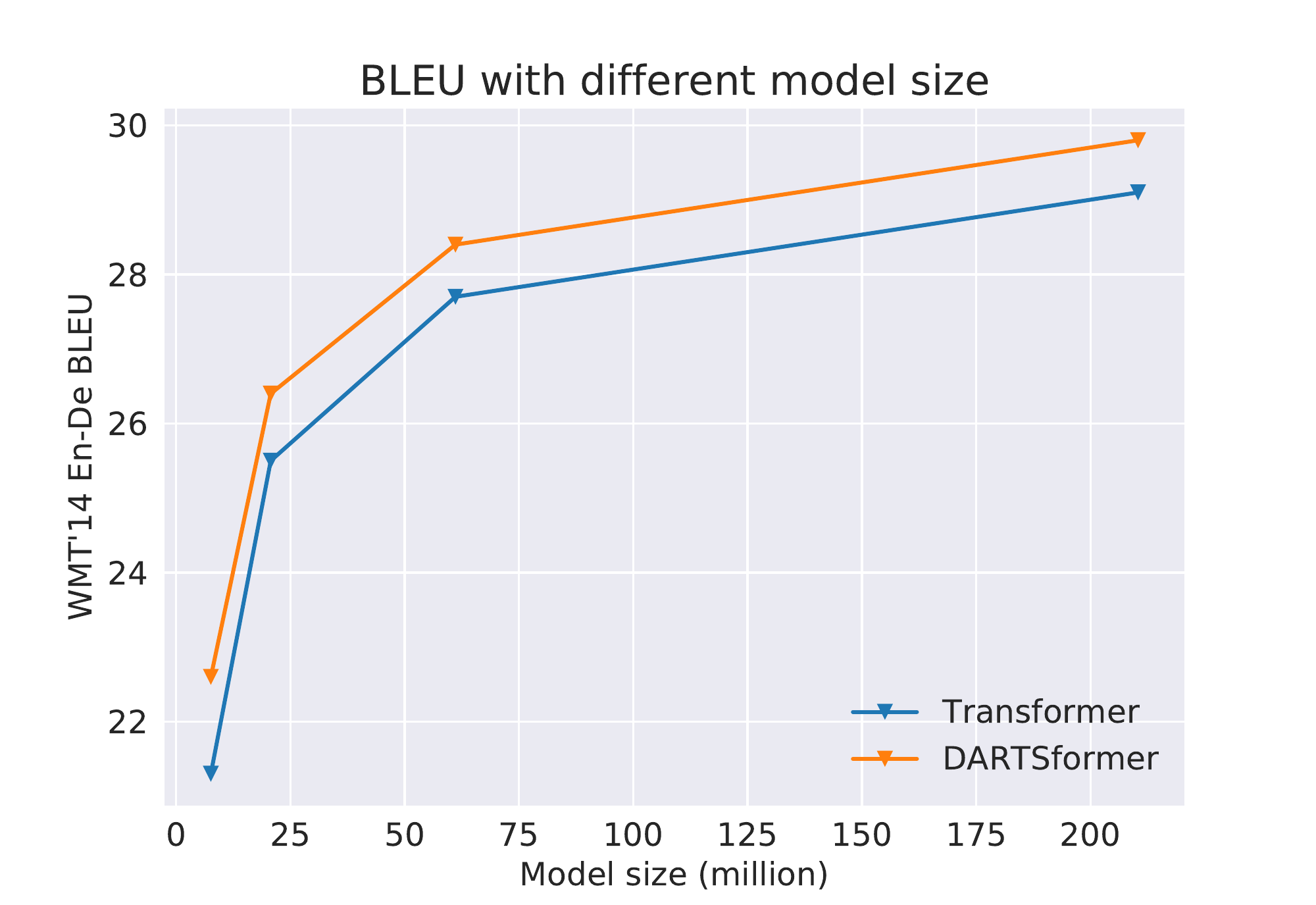}
    \caption{BLEU comparison between DARTSformer and standard Transformers with different model sizes.} \label{fig:BLEUCom}
\end{figure}

Figure \ref{fig:BLEUCom} shows the results for both architectures. DARTSformer performs better than standard Transformers at all sizes. The BLEU increase is [1.3/0.9/0.7/0.7] for [small/medium/base/big] models. An interesting fact is that the performance gap between two models tends to be smaller as we increase the model size, which is also observed in \citet{so2019evolved}. Based on this observation, DARTSformer is more pronounced for environments with resource limitations, such as mobile phones. A possible reason for the decreased performance gap at larger model sizes is that the effect of overfitting becomes more important. We expect that some data augmentation skills \citep{sennrich2015improving,edunov2018understanding, qu2020coda} might be of help.

\subsection{The Impact of Search Hidden Size}
The main motivation for our presented method is that we want to search in a large hidden size to reduce the performance gap between searching and re-training. However, whether this gap exists needs rigorous validation. Otherwise, it would suffice to instead use a small hidden size $d$ in architecture search, and then increase $d$ after search for training the actual model. We experiment with 4 search hidden sizes, namely, $e=128, d=120$ (tiny), $e=128, d=240$ (small), $e=256, d=480$ (medium), $e=256, d=960$ (DARTSformer). $e$ is the word embedding size and $d$ is the hidden size as described in Section \ref{sect:SearchConf}. After obtaining the searched model, we set the model size to $e=256, d=960$, and re-train it on WMT'14 En-De. 

The results are summarized in Table \ref{table:SmallSearchSizeRes}, which clearly shows that the translation quality is improving as the search hidden size gets larger. Also, note that when searching with tiny, small and medium settings, the final BLEU scores fall behind that of standard transformers. We argue that if one wants to evaluate the searched model in large model sizes, it is important to search with large hidden sizes. Further more, we directly apply DARTS with standard transformer as the backbone model. We set $e=320, d=320$. A larger search hidden size often causes memory failure due to the storage of many intermediate hidden states. As shown in Table \ref{table:SmallSearchSizeRes}, We can see that searching with a small hidden size yields no performance gain on the standard transformer. 

\begin{table}[t]
    \centering
    \begin{tabular}{lccccc}
        \toprule[1.5pt]
         \textbf{Search Settings} & \textbf{$e$} & \textbf{$d$} & \textbf{BLEU} \\
        \midrule
        Tiny & 128 & 120 & 24.2 \\
        Small & 128 & 240 & 26.3 \\
        Medium & 256 & 480 &  27.5 \\
        DARTSformer & 256 & 960 & \textbf{28.4} \\
         \midrule
        DARTS + Transformer & 320 & 320 & 27.7 \\
        Transformer & - & - & 27.7 \\
        \bottomrule[1.5pt]
    \end{tabular}
    \caption{BLEU scores of DARTS with different search hidden sizes on WMT'14 En-De test set. All searched architectures are re-trained with a parameter size similar to DARTSformer.}
    \label{table:SmallSearchSizeRes}
\end{table}

\section{Related Work}
\paragraph{Architecture Search} The field of neural architecture search (NAS) has seen advances in recent years. In the early stage, researchers focus on the reinforcement learning-based approaches \citep{baker2016designing,zoph2016neural,cai2018path,zhong2018practical} and evolution-based approaches \citep{liu2017hierarchical,real2017large,miikkulainen2019evolving,so2019evolved,wangetal2020hat}. These methods can produce architectures that outperform human-designed ones \citep{zoph2018carning,real2019regularized}. However, the computational cost is almost unbearable since it needs to fully train and evaluate every candidate network found in the search process. Weight sharing \citep{brock2017smash,pham2018efficient} is a practical solution where a super network is trained, and its sub-graphs form the search space. \citet{liu2018darts} proposed DARTS to use search parameters together with a super network, which allows searching with gradient descent. Gradient-based methods \citep{cai2018proxylessnas,xie2018snas,chen2019progressive,xu2019pc,yao2020efficient} attracts researchers' attention since it is computationally efficient and easy to implement. We base our method on DARTS and take one step further to reduce the memory consumption of training the super network. Another recent trend is the one-stage NAS \citep{cai2019once,mei2019atomnas,hu2020dsnas,yang2020ista}. Many NAS algorithms are in two stages. In the first stage, one searches for a good candidate network. In the second stage, the resulting network is re-initialized and re-trained. One-stage NAS tries to search and optimize the network weights simultaneously. After searching, one can have a ready-to-run network. We use a simple one-stage NAS algorithm \citep{guo2020single} as a baseline in Section \ref{sect:CompSetup}.
\paragraph{Reversible networks} The idea of reversible networks is first introduced by RevNets \citep{gomez2017reversible}. Later on, \citet{irevnet,Chang2018ReversibleAF,pmlrv97behrmann19a} invented different reversible architectures based on the ResNet \citep{he2016deep}. \citet{revrnn2018} extended RevNets to the recurrent network, which is particularly memory-efficient. \citet{bai2019deep,bai2020multiscale} conducted experiments with reversible Transformers by fixed point iteration. \citet{kitaev2020reformer} combined local sensitive hashing attention with reversible transformers to save memory in training with long sequences. An important application of reversible networks is the flow-based models \citep{NIPS20188224,pmlrv80huang18d,tran2019discrete}. For sequence tasks, \citet{maetal2019flowseq} achieved success in non-autoregressive machine translation.

\section{Conclusion}
We have proposed a memory-efficient differentiable architecture search (DARTS) method on sequence-to-sequence tasks. In particular, we have first devised a multi-split reversible network whose intermediate layer outputs can be reconstructed from top to bottom by the last layer's output. We have then combined this reversible network with DARTS and developed a backpropagation-with-reconstruction algorithm to significantly relieve the memory burden during the gradient-based architecture search process. We have validated the best searched architecture on three translation tasks.

Our method consistently outperforms standard Transformers. We can achieve the same BLEU score as the original big Transformer does with 69\% fewer parameters. At a large model size, we surpass Evolved Transformers with a search cost lower by an order of magnitude. Our method is generic to handle other architectures, and we plan to explore more tasks in the future.

\section*{Acknowledgments}
Yuekai Zhao and  Zhihua Zhang have been supported by the Beijing Natural Science Foundation (Z190001), National Key Research and Development Project of China (No. 2018AAA0101004), and Beijing Academy of Artificial Intelligence (BAAI).

\bibliography{DARTSformer}
\bibliographystyle{acl_natbib}

\end{document}